%% file: a.tex
\documentclass[conference]{IEEEtran}
\usepackage{cite}
\usepackage{amsmath,amssymb,amsfonts}
\usepackage{dsfont}
\usepackage{algorithmic}
\usepackage{graphicx}
\usepackage{textcomp}
\usepackage{xcolor}
\usepackage{url}

\usepackage[square,sort,comma,numbers]{natbib}
\usepackage{url}
\usepackage{multirow}
\usepackage{subfigure}
\usepackage[ruled,linesnumbered]{algorithm2e}

\begin{document}
	\sloppy
	
	\title{Deep Embedded K-Means Clustering}
	
	\author{\IEEEauthorblockN{Wengang Guo}
		\IEEEauthorblockA{\textit{Tongji University} \\
			Shanghai, China \\
			guowg@tongji.edu.cn}
		\and
		\IEEEauthorblockN{Kaiyan Lin$^\ast$}
		\IEEEauthorblockA{\textit{Tongji University} \\
			Shanghai, China \\
			ky.lin@163.com}
		\and
		\IEEEauthorblockN{Wei Ye$^\ast$}
		\IEEEauthorblockA{\textit{Tongji University} \\
			Shanghai, China \\
			yew@tongji.edu.cn}
	}
	


	
	\maketitle

	\begingroup\renewcommand\thefootnote{$^\ast$}
	\footnotetext{Corresponding author.}
	\endgroup
	
	\begin{abstract}
	Recently, deep clustering methods have gained momentum because of the high representational power of deep neural networks (DNNs) such as autoencoder. The key idea is that representation learning and clustering can reinforce each other: Good representations lead to good clustering while good clustering provides good supervisory signals to representation learning. Critical questions include: 1) How to optimize representation learning and clustering? 2) Should the reconstruction loss of autoencoder be considered always? In this paper, we propose DEKM (for Deep Embedded K-Means) to answer these two questions. Since the embedding space generated by autoencoder may have no obvious cluster structures, we propose to further transform the embedding space to a new space that reveals the cluster-structure information. This is achieved by an orthonormal transformation matrix, which contains the eigenvectors of the within-class scatter matrix of K-means. The eigenvalues indicate the importance of the eigenvectors' contributions to the cluster-structure information in the new space. Our goal is to increase the cluster-structure information. To this end, we discard the decoder and propose a greedy method to optimize the representation. Representation learning and clustering are alternately optimized by DEKM. Experimental results on the real-world datasets demonstrate that DEKM achieves state-of-the-art performance.
	\end{abstract}

	\IEEEpeerreviewmaketitle
	
	\input{sec_introduction}
	\input{sec_relatedwork}
	\input{sec_ourapproach}
	\input{sec_experiments}

	\input{sec_conclusion}
	
	\section*{Acknowledgment}
	
	The authors would like to thank anonymous reviewers for their constructive and helpful comments. This work was supported partially by the National Key Research and Development Program of China (Project No. 2020YFD1100603), the National Natural Science Foundation of China (NSFC) (Project No. 62176184) and the Fundamental Research Funds for the Central Universities.

	\bibliographystyle{plainnat} 
	\bibliography{reference}
	
\end{document}

%% file: sec_introduction.tex
\section{Introduction}
Clustering, an essential data exploratory analysis tool, has been widely studied. Numerous clustering methods have been proposed, e.g., K-means \cite{lloyd1982least}, Gaussian Mixture Models \cite{bishop2006pattern}, and spectral clustering \cite{von2007tutorial,shi2000normalized,ng2002spectral,ye2016fuse}. These traditional clustering methods have achieved promising results. However, when the dimension of the data is very high, these traditional methods become inefficient and ineffective because of the notorious problem of the curse of dimensionality.

To deal with this problem, people usually perform dimensionality reduction before clustering. 
The commonly used dimensionality reduction methods are Principle Component Analysis (PCA)~\cite{pearson1901liii}, Canonical Correlation Analysis (CCA)~\cite{andrew2013deep}, and Nonnegative Matrix Factorization (NMF)~\cite{lee1999learning}. But all these methods are linear and shallow models, whose representational abilities are limited. They cannot capture the complex and non-linear relations hidden in the data. Benefiting from the high representational power of deep neural networks (DNNs), autoencoder has been widely used as a dimensionality reduction method for clustering in recent years. Autoencoder can learn meaningful representations of the input data through an unsupervised manner. It consists of an encoder and a decoder.
\begin{figure}[!htb]
	\hspace*{\fill}
	\centering	
	\subfigure[Iter 0, (92.0, 79.9)]{\includegraphics[width=0.18\textwidth]{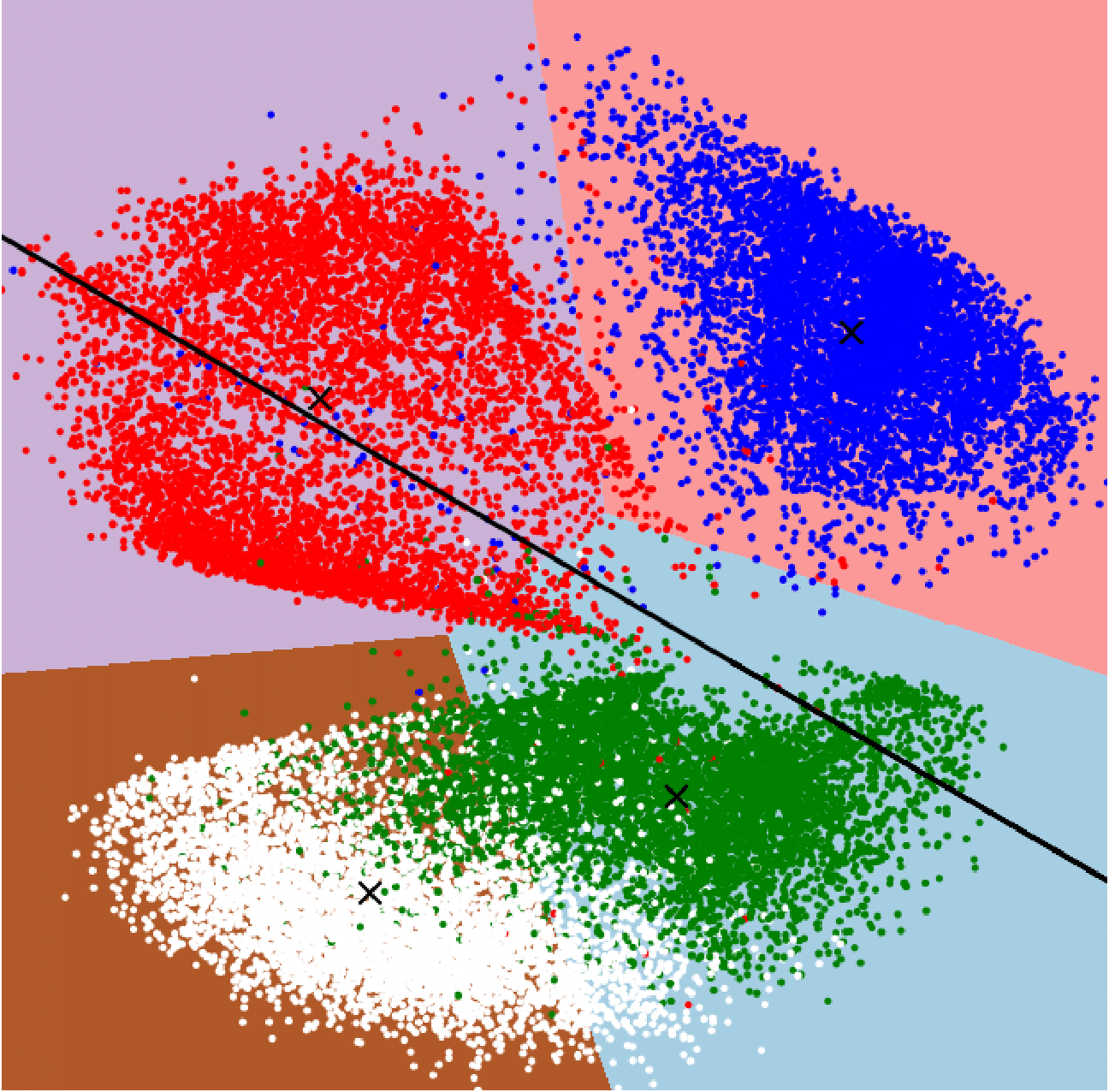}}
	\hfill
	\centering
	\subfigure[Iter 1, (94.6, 84.3)]{\includegraphics[width=0.18\textwidth]{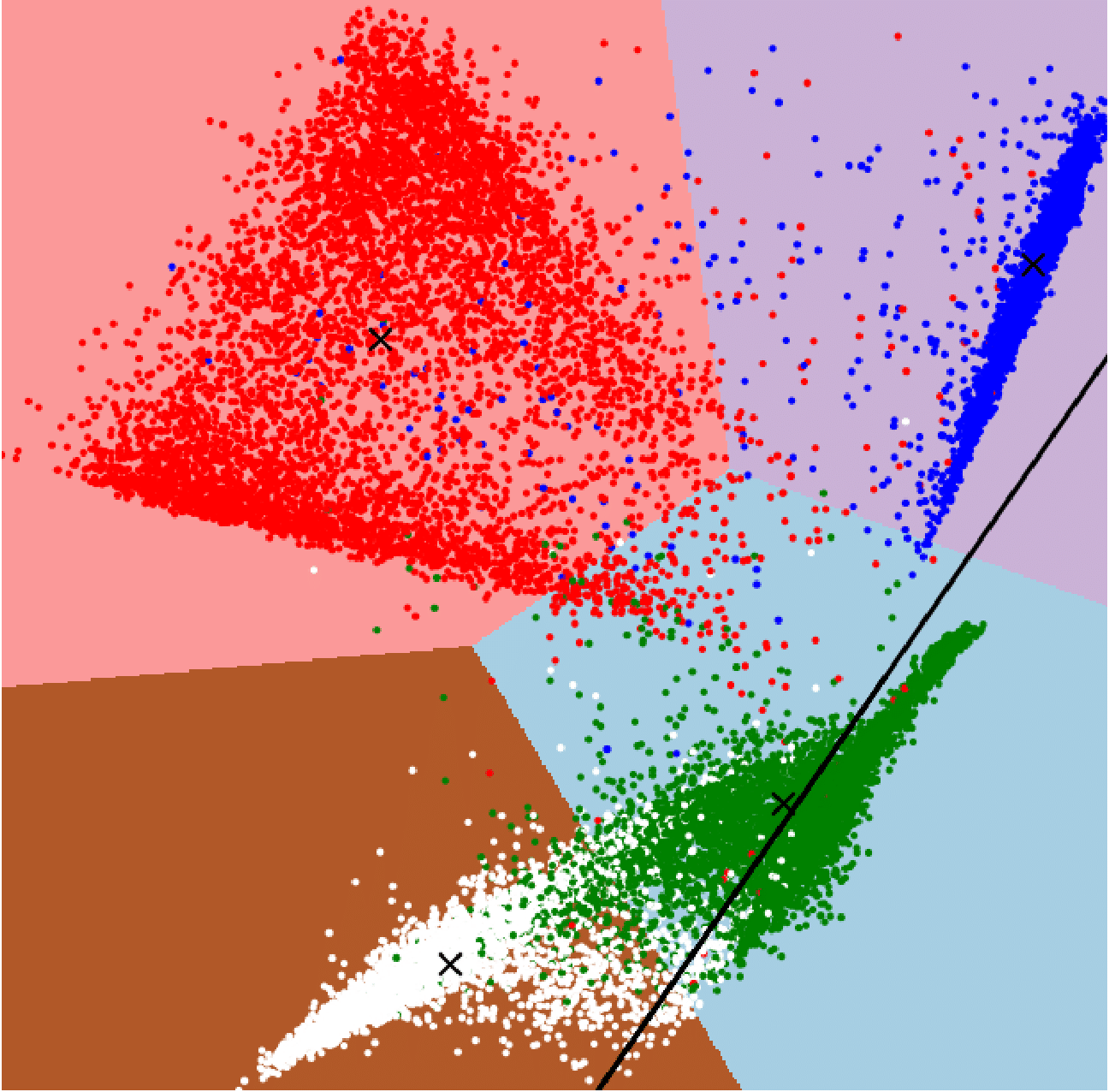}}
	\hspace*{\fill}
	
	\hspace*{\fill}
	\centering
	\subfigure[Iter 2, (94.8, 84.9)]{\includegraphics[width=0.18\textwidth]{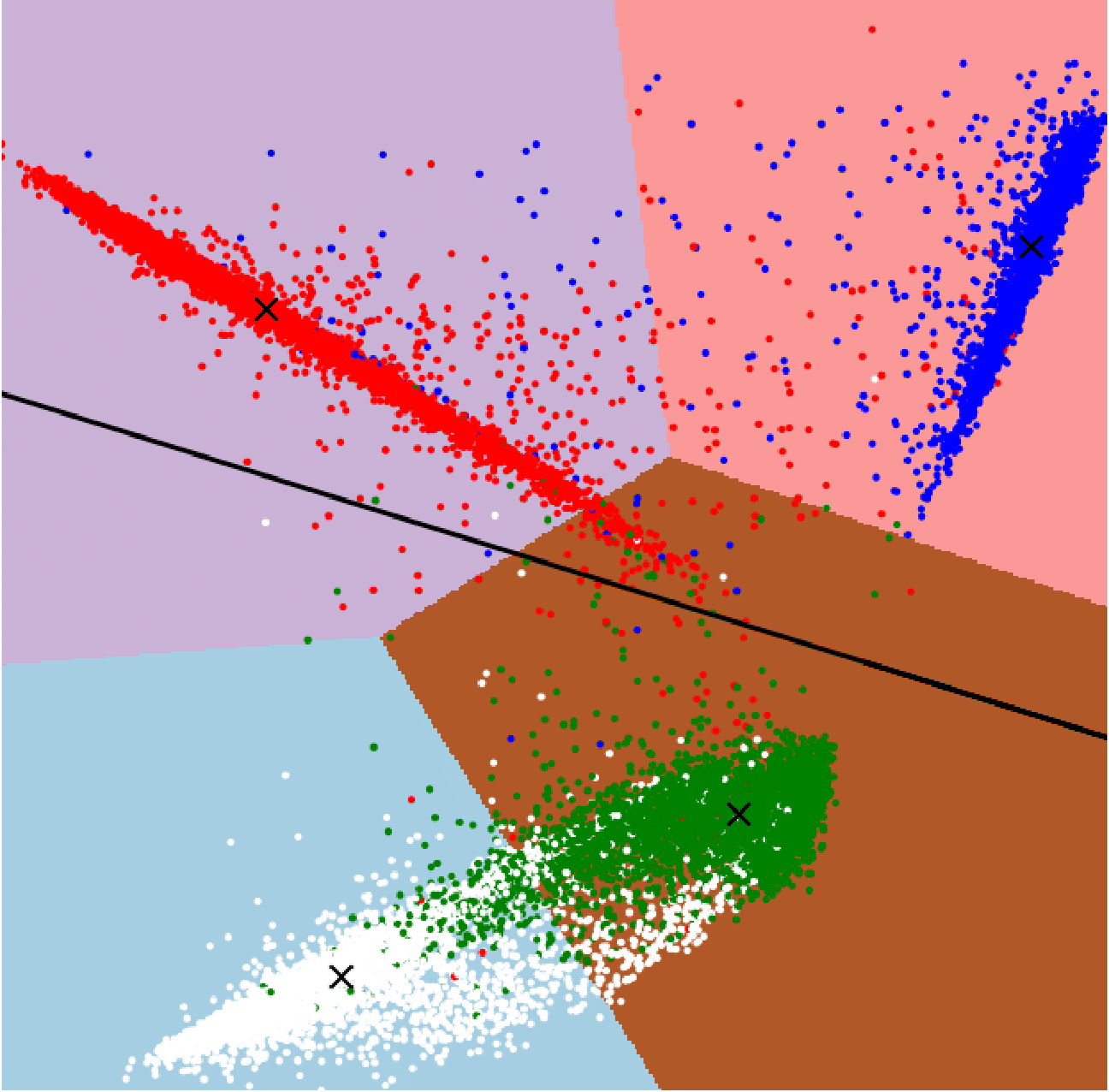}}
	\hfill
	\centering
	\subfigure[Iter 20, (97.0, 89.8)]{\includegraphics[width=0.18\textwidth]{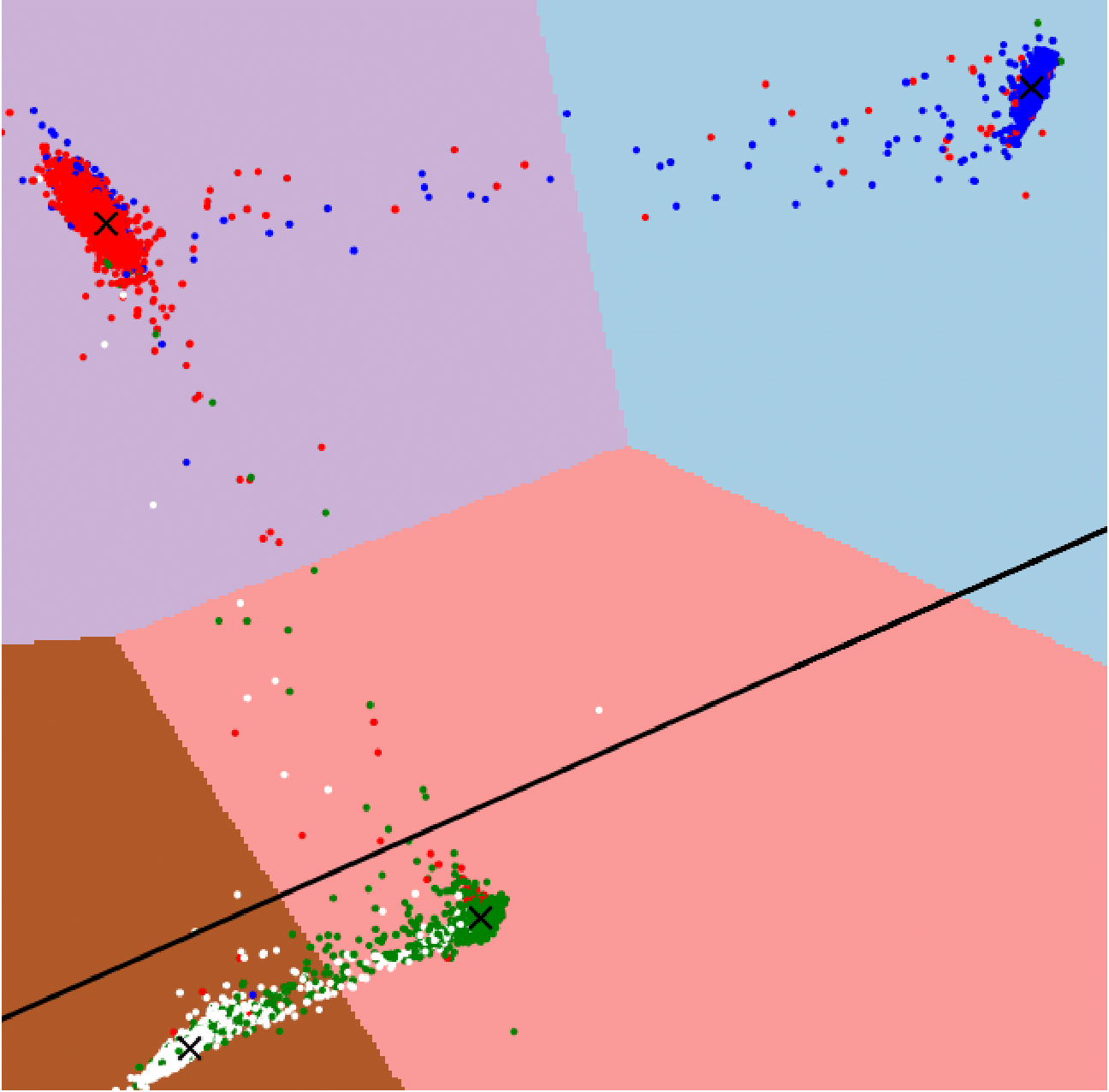}}
	\hspace*{\fill}
	
	\caption{Visualization of the representation learning and clustering by DEKM on a subset of MNIST dataset. The color of data points marks the ground truth and the color of the background block marks the clustering results. Data points falling into the same block belong to the same cluster. Cluster centroids are marked by black crosses. The black line shows the direction of the last eigenvector of the within-class scatter matrix of K-means. In this direction, the cluster-structure information is the lowest. We would like to optimize the representation to increase the cluster-structure information in this direction. We use a mini-batch updating strategy. Numbers in parentheses denote ACC and NMI, respectively.}
	\label{fig_2D}
\end{figure}
The encoder transforms the input data into a lower-dimensional space (the embedding space), and the decoder is responsible for reconstructing the input data from that embedding space.

A plethora of clustering methods~\cite{yang2016joint,li2018discriminatively,DEC_xie2016unsupervised,DEPICT_ghasedi2017deep,DCN_yang2017towards,IDEC_guo2017improved,DCEC_guo2017deep,shah2018deep,fard2020deep} based on autoencoder have been proposed. One of the most critical problems with deep clustering is how to design a proper clustering loss function. Methods such as DEC~\cite{DEC_xie2016unsupervised} and IDEC~\cite{IDEC_guo2017improved} minimize the Kullback-Leibler (KL) divergence between the cluster distribution and an auxiliary target distribution. The basic idea is to refine the clusters by learning from their high confidence assignments. However, the auxiliary target distribution is hard to choose for better clustering results. Other methods such as DCN~\cite{DCN_yang2017towards} and DKM~\cite{fard2020deep} combine the objective of K-means with that of autoencoder and jointly optimize them. However, the discrimination of clusters in the embedding space is not directly related to the reconstruction loss of auoencoder. Thus, in this paper, after using an autoencoder to generate an embedding space, we discard the decoder and do not optimize the reconstruction loss anymore. Because no matter what we do to the embddding space, we can separately train the decoder to reconstruct the input data from their embeddings.

Considering the simplicity of K-means, we extend it to a deep version, which is called DEKM that uses an autoencoder to generate an embedding space. Since this embedding space may not be discriminative for clustering, we propose to further transform this embedding space to a new space by an orthonormal transformation matrix, which consists of the eigenvectors of the within-class scatter matrix of K-means. These eigenvectors are sorted in ascending order with respect to their eigenvalues. In this new space, the cluster-structure information is revealed. And each eigenvalue indicates how much its corresponding eigenvector contributes to the quantity of the cluster-structure information in the new space. The last eigenvector contributes the least. To increase the cluster-structure information in the direction of the last eigenvector, we discard the decoder and propose a greedy method to optimize the representation. Optimizing representation is equivalent to minimizing the entropy. And this optimization is also in consistent with the loss of K-means. Inspired by the idea that ``\textit{good representations are beneficial to clustering and clustering provides supervisory signals to representation learning}''~\cite{yang2016joint}, we alternately optimize representation learning and clustering until some criterion is met. 

Figure~\ref{fig_2D} shows these two alternating processes on a subset of MNIST dataset, containing randomly selected four clusters. We set the units of the embedding layer to two for better inspection. Figure~\ref{fig_2D}(a)--(d) show the data representations and clustering results in the embedding space during these two alternating processes. Figure~\ref{fig_2D}(a) shows the initial embedding space generated by an autoencoder. The direction where the black line lies denotes the second eigenvector of the scatter matrix of K-means. The cluster-structure information in this direction is the lowest. To increase the cluster-structure information in this direction, we move data points closer to their centroids. Note that in this process, we use a mini-batch of data to optimize the representation, which makes only some data points closer to their centroids. Compared with Figure~\ref{fig_2D}(a), we can see from Figure~\ref{fig_2D}(b) that the data points in the red cluster do not move closer to their centroids, while the data points in the other three clusters do. We find in the experiments that this mini-batch updating strategy outperforms the full-batch updating strategy, which moves all the data points closer to their centroids.

Our contributions can be summarized as follows:
\begin{itemize}
	\item We propose to transform the embedding space generated by an autoencoder to a new space using an orthonormal transformation matrix, which contains the eigenvectors of the within-class scatter matrix of K-means. The eigenvalues indicate the importance of their corresponding eigenvectors' contributions to the cluster-structure information in the new space.
	\item We develop a greedy method to optimize the representation so that the cluster-structure information in the new space is increased.
	\item We alternately optimize representation learning and clustering.
	\item We show the effectiveness of DEKM by carrying out extensive comparative studies with state-of-the-art methods on various real-world datasets.
\end{itemize}


%% file: sec_relatedwork.tex
\section{Related Work}

\subsection{K-means and Its Variants}

K-means is one of the most fundamental clustering methods. It is usually used as one of the building blocks of many advanced clustering methods such as spectral clustering~\cite{von2007tutorial,shi2000normalized,ng2002spectral,ye2016fuse}. K-means has inspired many extentions. For example, the basic idea of \cite{jain1988algorithms} is to replace the mean with the median. K-means++ \cite{arthur2006k} improves the selection of the initial centroids, which are based on their proportional distance to the previous selected centroids. SubKmeans \cite{mautz2017towards} assumes that the input space can be split into two independent subspaces, i.e., the clustering subspace and the noise subspace. The former subspace contains only the cluster-structure information and the latter subspace contains only the noise information. SubKmeans performs clustering in the clustering subspace. Nr-Kmeans~\cite{mautz2018discovering,mautz2020non} finds non-redundant K-means clusterings in multiple mutually orthogonal subspaces via an orthogonal transformation matrix. Fuzzy-c-means~\cite{dunn1973fuzzy} assigns each data point proportionally to multiple clusters. It relaxes the hard cluster assignments of K-means to soft cluster assignments. Mini-batch K-means~\cite{sculley2010web} extends K-means to the scenario of user-facing web applications. Mini-batch K-means can be used in the deep learning frameworks because it supports the online Stochastic Gradient Descent (SGD). 

For the real-world datasets, the number of clusters is unknown. To solve this problem, researchers propose to automatically find the proper number of clusters. X-means~\cite{pelleg2000x} uses the Bayesian Information Criterion (BIC) or the Akaike Information Criterion (AIC) as measures to evaluate the clustering results under different cluster number $k$. 
G-means~\cite{hamerly2004learning} assumes that each cluster follows a Gaussian distribution. It runs K-means with increasing $k$ hierarchically until the statistical test accepts that the clusters follow Gaussian distributions. PG-means~\cite{feng2006pg} first constructs an one-dimensional projection of the dataset and the learned model. Then the model fitness is evaluated in the projected space. It is able to discover a proper number of Gaussian clusters. Dip-means~\cite{kalogeratos2012dip} assumes that each cluster follows a unimodal distribution. It first computes the pairwise distance between a data point and the other data points. Then, it applies a univariate statistical hypothesis test~\cite{hartigan1985dip} (called Hartigans' dip-test) for unimodality on the distributions of distances to find unimodal clusters and the proper cluster number. 

\subsection{Deep Clustering}

Since shallow clustering models suffer from non-linearity of real-world data, they cannot perform well. Deep clustering models use deep neural networks (DNNs) with stronger non-linear representational capabilities to extract features, which achieve better clustering performances. Earlier deep clustering methods~\cite{ding2004k,trigeorgis2014deep} perform representation learning and clustering sequentially. Recent studies~\cite{yang2016joint, li2018discriminatively,DEC_xie2016unsupervised,DEPICT_ghasedi2017deep,DCN_yang2017towards,IDEC_guo2017improved,DCEC_guo2017deep,shah2018deep,fard2020deep} have shown that jointly performing representation learning and clustering yields better performance. 


JULE~\cite{yang2016joint} proposes a recurrent framework for jointly unsupervised learning of deep representations and clustering. During the optimization procedure, clustering is conducted in the forward pass and representation learning is performed in the backward pass. DCC~\cite{shah2018deep} jointly performs non-linear dimensionality reduction and clustering. The clustering process includes the optimization of the autoencoder. Since the objective function is continuous without the discrete cluster assignment, it can be solved by standard gradient-based methods. DEC~\cite{DEC_xie2016unsupervised} pre-trains an autoencoder with reconstruction loss and performs clustering to obtain the soft clustering assignment of each data point. Then, an auxiliary target distribution is derived from the current soft cluster assignments. Finally, it iteratively refines clustering by minimizing the Kullback-Leibler (KL) divergence between the soft assignments and the auxiliary target distribution. DCN~\cite{DCN_yang2017towards} combines the objective of K-means with that of  autoencoder to find a ``K-means-friendly'' space. The cluster assignments of DCN are not soft (probability) like that in DEC, but strict (discrete), which limit the direct use of gradient-based SGD solvers. DCN refines clustering by alternately optimizing the objectives of autoencoder and K-means. DEPICT~\cite{DEPICT_ghasedi2017deep} consists of two parts, i.e., a convolutional autoencoder for learning the embedding space and a multinomial logistic regression layer functioning as a discriminative clustering model. For jointly learning the embedding space and clustering, DEPICT employs an alternating approach to optimize a unified objective function. IDEC~\cite{IDEC_guo2017improved} combines an under-complete autoencoder with DEC. The under-complete autoencoder not only learns the embedding space, but also preserves the local structure of data. Like IDEC, DCEC \cite{DCEC_guo2017deep} combines a convolutional autoencoder with DEC. DKM~\cite{fard2020deep} proposes a new approach for jointly clustering with K-means and learning representations. The K-means objective is considered as a limit of a differentiable function, so that the representation learning and clustering can be optimized by the simple stochastic gradient descent. RED-KC (for Robust Embedded Deep K-means Clustering)~\cite{zhang2019robust} uses the $\delta$-norm metric to constrain the feature mapping of the autoencoder so that data embeddings are more conductive to the robust K-means clustering.

Our proposed DEKM is also a method that jointly performs representation learning and clustering. Like DCEC, DEKM first uses an autoencoder to find the embedding space. Then, it discards the decoder and optimizes the representation for better clustering. The representation optimization of DEKM is different from that of DCEC, which does not optimize the Kullback-Leibler (KL) divergence between the cluster distribution and an auxiliary target distribution. Instead, DEKM optimizes the representation by reducing its entropy.

%% file: sec_ourapproach.tex
\section{Deep Embedded K-Means}

We assume that the cluster structures exist in a lower-dimensional subspace. Instead of clustering directly in the original space, DEKM uses autoencoder to transform the original space into an embedding space to reduce the dimensionality before clustering. DEKM alternately optimizes representation learning and clustering. DEKM has three steps: (1) generating an embedding space by an autoencoder, (2) detecting clusters in the embedding space by K-means, and (3) optimizing the representation to increase the cluster-structure information. The last two steps are alternately optimized to generate better embedding space and clustering results. Table~\ref{tab_symbol} shows the used symbols and their corresponding interpretations in this paper.

\begin{table}[htbp]
	\caption{Symbols and Interpretations}
	\begin{center}
		\begin{tabular}{|l|l|}
			\hline
			Symbol & Interpretation \\ \hline
			$d \in \mathbb{N}$ & The dimension of input space \\ 
			$e \in \mathbb{N}$ & The dimension of embedding space \\ 
			$n \in \mathbb{N}$ & The number of data points \\ 
			$k \in \mathbb{N}$ & The number of clusters \\ \hline
			$\mathbf{x} \in \mathbb{R}^{d\times 1}$ & A data point \\ 
			$\boldsymbol{\mu} \in \mathbb{R}^{e\times 1}$ & A cluster centroid \\ 
			$\mathbf{h} \in \mathbb{R}^{e\times 1}$ & A data point embedding \\ \hline
			$\mathbf{X} = \left[ \mathbf{x}_1;\ldots;\mathbf{x}_n\right] $     & Data matrix   \\ 
			$\mathbf{I}\in \mathbb{R}^{e\times e}$     & Identity matrix  \\
			$\mathbf{V} \in \mathbb{R}^{e\times e}$      & Orthonormal transformation matrix \\
			$\mathbf{H} = \left[ \mathbf{h}_1;\ldots;\mathbf{h}_n\right] $     & Data embeddings  \\ \hline
			$\mathcal{U}=\left\{\boldsymbol{\mu}_i\right\}_{i = 1}^k$      & Set of all cluster centroids   \\ 
			$\mathcal{C}=\left\{\mathcal{C}_1,\ldots,\mathcal{C}_k\right\}$      &Set of clusters \\ \hline
			$f(\cdot)$	&     Encoder       \\    
			$g(\cdot)$	&    Decoder           \\ \hline
		\end{tabular}
	\end{center}
	\label{tab_symbol}
\end{table}

\subsection{Generating an Embedding Space}
\label{sec:step1}

Autoencoder is a type of deep neural networks (DNNs) that can learn low-dimensional representations of the input data in an unsupervised way. It consists of an encoder and a decoder. The encoder $f(\cdot)$ transforms the input data into a lower-dimensional space (the embedding space), and the decoder $g(\cdot)$ reconstructs the input data from the embedding space. Autoencoder is trained to minimize the reconstruction loss, such as the least-squares error:
\begin{equation}
\label{eqn:ae}
	\begin{aligned}
		\min L_1 &= \sum_{i=1}^n\left\| \mathbf{x}_i-\mathbf{\widehat{x}}_i\right\|^2\\
		&= \sum_{i=1}^n\left\| \mathbf{x}_i - g\left(f(\mathbf{x}_i)\right) \right\|^2
	\end{aligned}
\end{equation}
where $\mathbf{x}_i$ is the $i$-th data point, $f(\mathbf{x}_i)$ is the output of the encoder $f(\cdot)$, and $\mathbf{\widehat{x}}_i$ is the reconstructed output of the decoder $g(\cdot)$. The dimension of the embedding space usually is set to a number that is much smaller than that of the original space. This not only mitigates the curse of dimensionality, but also helps to avoid the trivial solution of autoencoder, of which both $f(\cdot)$ and $g(\cdot)$ equal to the identity matrix. 



\subsection{Detecting Clusters}\label{sec:step2}

In the above section, we train the autoencoder using the least-squares error loss to generate an embedding space $\mathbf{H}=f(\mathbf{X})$, without considering the characteristics of the embedding space. This embedding space may not contain any cluster structures. DCN~\cite{DCN_yang2017towards} combines the objective function of the autoencoder with that of K-means and optimizes them alternately. DCN wants to find a ``K-means-friendly'' subspace. However, the relative importance parameter between the two objective functions is hard to set. In addition, this paradigm is difficult to generate a ``K-means-friendly'' subspace because of the reconstruction loss of the autoencoder. In the optimization procedure, the reconstruction loss of the autoencoder should not be used anymore. The reason is that whatever we modify on the encoder, we can still train the decoder to make Equation~(\ref{eqn:ae}) minimized.


We use K-means \cite{lloyd1982least} to find a partition $\left\{ \mathcal{C}_i \right\}_{i = 1}^k$ of the data points in the embedding space $\mathbf{H}$. Its objective function is as follows:
\begin{equation}
\label{eqn:kmeans}
\min L_2 = \sum_{i = 1}^k \sum_{\mathbf{h} \in \mathcal{C}_i} \left\| \mathbf{h} - \boldsymbol{\mu}_i \right\|^2\\
\end{equation}
where $\mathbf{h}$ is a data point in the embedding space, $k$ is the number of clusters, $\mathcal{C}_i$ denotes the set of data points assigned to the $i$-th cluster, $\boldsymbol{\mu}_i=\frac{1}{\left| \mathcal{C}_i\right| }\sum_{\mathbf{h} \in \mathcal{C}_i}\mathbf{h}$ denotes the centroid of the $i$-th cluster. 

To reveal the cluster structures in the embedding space, we propose to transform the embedding space $\mathbf{H}$ to a new space by an orthonormal transformation matrix $\mathbf{V}$. In the new space $\mathbf{Y}=\mathbf{V}\mathbf{H}$, Equation (\ref{eqn:kmeans}) becomes the following:
\begin{equation}
\label{eqn:transform}
\begin{split}
\min L_3 &= \sum_{i = 1}^k \sum_{\mathbf{h} \in \mathcal{C}_i} \left\| \mathbf{V}\mathbf{h} - \mathbf{V}\boldsymbol{\mu}_i \right\|^2\\
&=	\sum_{i = 1}^k \sum_{\mathbf{h} \in \mathcal{C}_i} \left(\mathbf{V}\mathbf{h} - \mathbf{V}\boldsymbol{\mu}_i\right)^\intercal\left(\mathbf{V}\mathbf{h} - \mathbf{V}\boldsymbol{\mu}_i\right)\\
&=\sum_{i = 1}^k \sum_{\mathbf{h} \in \mathcal{C}_i}\left(\mathbf{h} - \boldsymbol{\mu}_i\right)^\intercal\mathbf{V}^\intercal\mathbf{V}\left(\mathbf{h} - \boldsymbol{\mu}_i\right)\\
&=\sum_{i = 1}^k \sum_{\mathbf{h} \in \mathcal{C}_i}\mbox{Trace}\left(\left(\mathbf{h} - \boldsymbol{\mu}_i\right)^\intercal\mathbf{V}^\intercal\mathbf{V}\left(\mathbf{h} - \boldsymbol{\mu}_i\right)\right)
\end{split}
\end{equation}
where we use the trace-trick in the last step, because a scalar can also be considered as a matrix of size $1\times 1$. Since $\mathbf{V}^\intercal\mathbf{V}=\mathbf{I}$, minimizing Equation (\ref{eqn:transform}) is equivalent to minimizing Equation (\ref{eqn:kmeans}).

The above equation can be further written as:
\begin{equation}
\label{eqn:withclass}
\begin{split}
\min L_3 &= \mbox{Trace}\left(\mathbf{V}\left[ \sum_{i = 1}^k \sum_{\mathbf{h} \in \mathcal{C}_i}\left(\mathbf{h} - \boldsymbol{\mu}_i\right)\left(\mathbf{h} - \boldsymbol{\mu}_i\right)^\intercal\right] \mathbf{V}^\intercal\right)\\
&=\mbox{Trace}\left(\mathbf{V}\mathbf{S}_{w}\mathbf{V}^\intercal\right)
\end{split}
\end{equation}
where $\mathbf{S}_{w}=\sum_{i = 1}^k \sum_{\mathbf{h} \in \mathcal{C}_i}\left(\mathbf{h} - \boldsymbol{\mu}_i\right)\left(\mathbf{h} - \boldsymbol{\mu}_i\right)^\intercal$ is the within-class scatter matrix of K-means.

Since $\mathbf{V}$ is an orthonormal matrix, minimizing Equation~(\ref{eqn:withclass}) is a standard trace minimization problem. A version of the Rayleigh-Ritz theorem~\cite{lutkepohl1997handbook} indicates that the solution $\mathbf{V}$ contains the eigenvectors of $\mathbf{S}_w$ with ascending eigenvalues. The eigenvalues indicate the importance of the eigenvectors' contributions to the cluster structures in the transformed space $\mathbf{Y}=\mathbf{V}\mathbf{H}$. The smaller the eigenvalue, the more important its corresponding eigenvector contributes to the cluster structures in the transformed space $\mathbf{Y}$. We should note that $\mathbf{S}_w$ is symmetric and therefore it is orthogonally diagonalizable. So it is feasible to find the orthonormal matrix $\mathbf{V}$.

\subsection{Optimizing Representation}\label{sec:step3}

As discussed above, minimizing Equation (\ref{eqn:transform}) equals to minimizing Equation (\ref{eqn:kmeans}). We can first perform K-means in the embedding space $\mathbf{H}$ to get $\mathbf{S}_{w}$, and then eigendecompose $\mathbf{S}_{w}$ to get $\mathbf{V}$. Finally, we transform the embedding space to a new space $\mathbf{Y}$ that reveals the cluster-structure information. We also know the importance of each dimension of $\mathbf{Y}$ in terms of the cluster-structure information, i.e., the last dimension has the least cluster-structure information. 

We can rewrite Equation (\ref{eqn:transform}) as follows:
\begin{equation}
\label{eqn:transform1}
\min L_3 = \sum_{i = 1}^k \sum_{\mathbf{y} \in \mathcal{C}_i} \left\| \mathbf{y} - \mathbf{m}_i \right\|^2
\end{equation}
where $\mathbf{y}=\mathbf{V}\mathbf{h}$ and $\mathbf{m}_i=\mathbf{V}\boldsymbol{\mu}_i$.

\begin{figure*}[!htb]
	\centering	
	\subfigure[A uniform distribution. The entropy is 5.992 bits.]{\includegraphics[width=0.3\textwidth]{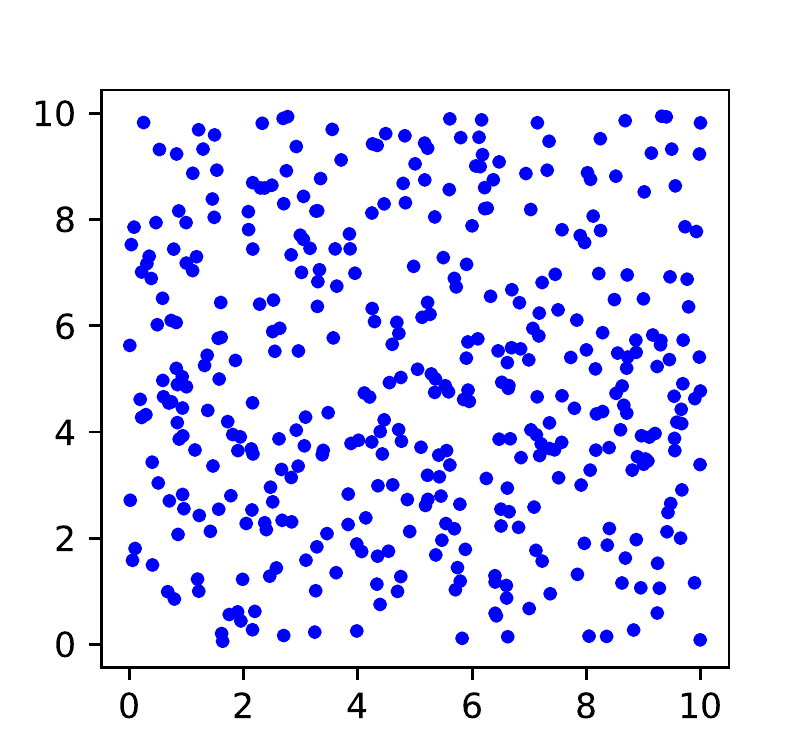}}
	\hspace{1em}
	\subfigure[Two Gaussian distributions with large variance. The entropy is 1.419 bits.]{\includegraphics[width=0.3\textwidth]{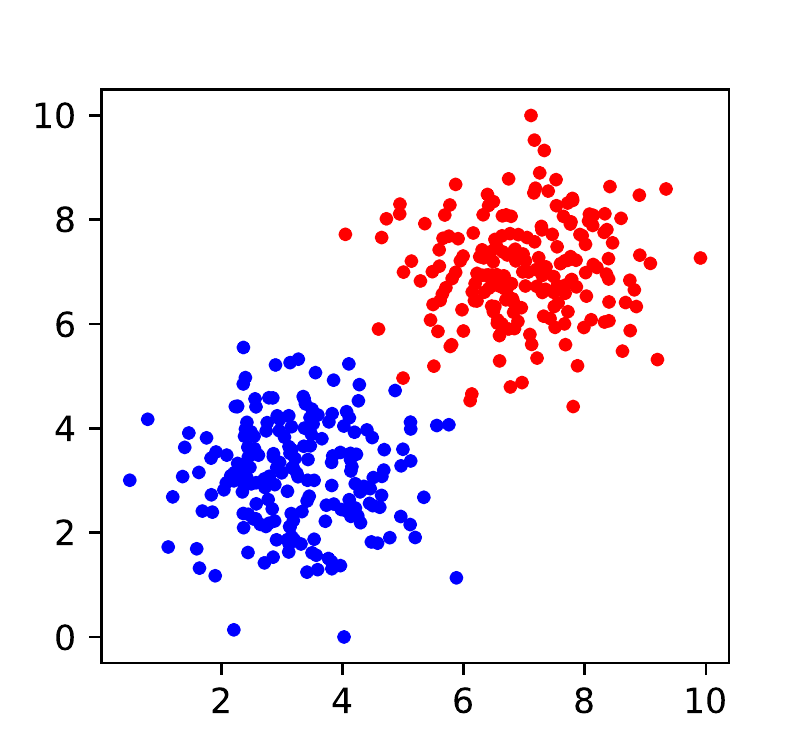}}
	\hspace{1em}
	\subfigure[Two Gaussian distributions with low variance. The entropy is 0.033 bits.]{\includegraphics[width=0.3\textwidth]{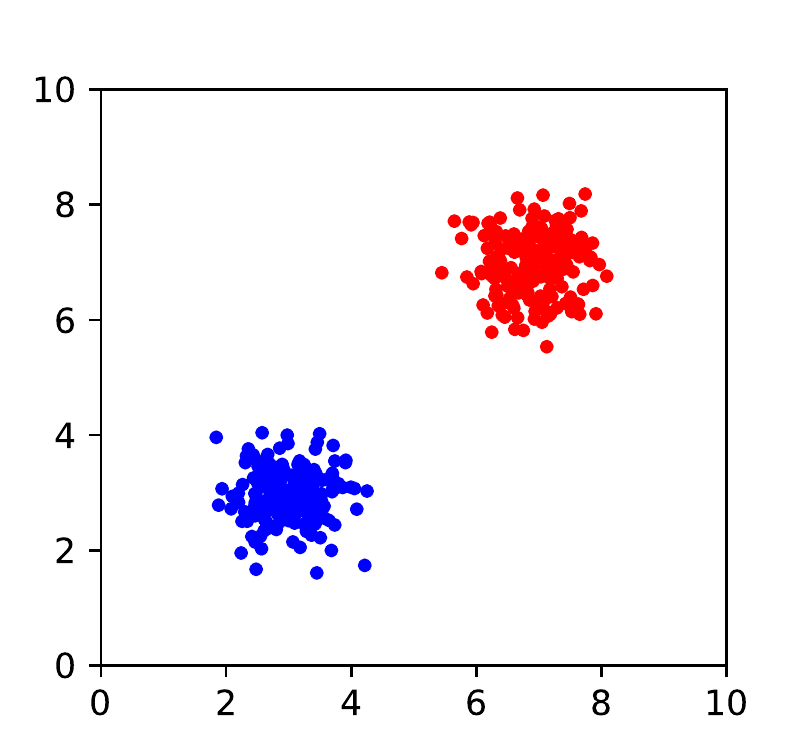}}
	\caption{Demonstration of the relationships between entropy and cluster-structure information. The lower the entropy of data, the higher the cluster-structure information it contains. Entropy decreases from (a) to (c), while cluster-structure information increases from (a) to (c).}
	\label{fig:prove}
\end{figure*}

Now the question is how to optimize the representation to improve the cluster-structure information in $\mathbf{Y}$. In this paper, we measure the cluster-structure information by \textbf{entropy}. \textbf{The lower the entropy of data, the higher the cluster-structure information it contains}. For example, Figure~\ref{fig:prove}(a) shows 400 data points that follow a uniform distribution in the range $[0, 10]$. The entropy is $-\sum_{i=1}^{400}\frac{1}{400}\ln\left(\frac{1}{400}\right)=5.992$ bits. Compared with Figure~\ref{fig:prove}(a), there are two Gaussian clusters in Figure~\ref{fig:prove}(b), each of which has 200 data points. The two Gaussian clusters have the same variance of one. The two Gaussian clusters in Figure~\ref{fig:prove}(c) have the same variance of 0.5, each of which also has 200 data points. Note that the entropy of a $d$-dimensional Gaussian distribution is $\frac{d}{2}\ln\left(2\pi e\sigma_1^2\sigma_2^2\ldots\sigma_n^2\right)^{1/d}$ as shown in~\cite{osti_5259367}. Thus, the entropy of data in Figure~\ref{fig:prove}(b) is 1.419 bits, and that of data in Figure~\ref{fig:prove}(c) is 0.033 bits. From Figure~\ref{fig:prove}, we have two observations: 1) Compared with data that has no cluster-structure information, data that contains the cluster-structure information has lower entropy. 2) Reducing the variance of a Gaussian cluster will reduce its entropy. Also note that the clusters found by K-means follow isotropic Gaussian distributions. Based on these, we propose a new objective to optimize the representation to increase the cluster-structure information. Specifically, we would like to \textbf{move the data points in clusters found by K-means closer to their centroids}. This strategy is also in consistent with Equation (\ref{eqn:kmeans}), which is further minimized.


Instead of moving data points closer to their respective centroids in each dimension of $\mathbf{Y}$, we propose a greedy method that only moves data points closer to their respective centroids in the last dimension of $\mathbf{Y}$. In this way, the representation will be easily optimized to increase the cluster-structure information. We found in our experiments that this greedy method performs the best. The details of the greedy method is as follows: We first replicate $\mathbf{y}$ as $\mathbf{y}'$ and then replace the last dimension of $\mathbf{y}'$ with the last dimension of $\mathbf{m}_i$. Finally, The objective function is defined as follows:
\begin{equation}
\label{eqn:finetune}
\min L_4 = \sum_{i = 1}^k \sum_{\mathbf{y} \in \mathcal{C}_i} \left\| \mathbf{y} - \mathbf{y}' \right\|^2
\end{equation}

We do not use a full-batch updating strategy that moves all the data points closer to their centroids in the last dimension. Instead, we use a mini-batch updating strategy that moves only some mini-batches of data points closer to their centroids. We find that the mini-batch updating strategy is superior to the full-batch updating strategy. After optimizing the representation, we have a new embedding space $\mathbf{H}$. Then, we perform K-means again to find clusters and their respective centroids. We alternately repeat the second and third steps until some criterion is met, such as the predefined number of iterations or there are less than 0.1\% of samples that change their clustering assignments between two consecutive iterations. The pseudo-code of DEKM is shown in Algorithm~\ref{algo_1}. Line 1 uses Equation (\ref{eqn:ae}) to train the autoencoder. Line 3 uses the encoder to generate an embedding space $\mathbf{H}=f(\mathbf{X})$. Then in the embedding space, line 4 performs K-means to find clusters. Lines 5--6 compute the within-class scatter matrix $\mathbf{S}_{w}$ and eigendecompose it to get the orthonormal transformation matrix $\mathbf{V}$. Line 7 uses Equation~(\ref{eqn:finetune}) to optimize the representation. We repeat the process for a number $Iter$ of iterations and return the final cluster set $\mathcal{C}$.


\begin{algorithm}
	\label{algo_1}
	\caption{DEKM}
	\SetKw{KwKmeans}{Kmeans}
	\SetKw{KwEig}{Eig}
	\KwIn{ Data matrix $\mathbf{X}$, the number of clusters $k$, and the maximum number of iterations $Iter$}
	\KwOut{Cluster set $\mathcal{C}=\left\{\mathcal{C}_1,\ldots,\mathcal{C}_k\right\}$}
	Using Equation (\ref{eqn:ae}) to train the autoencoder\;
	\For{$i \leftarrow 1$ \KwTo $Iter$ }{
	Generate an embedding space by the encoder $\mathbf{H}=f(\mathbf{X})$\;
	$\left(\mathcal{C}, \mathcal{U}\right)\leftarrow \KwKmeans \left(\mathbf{H}, k\right)$\;
	Compute the within-class scatter matrix $\mathbf{S}_{w}\leftarrow\sum_{i = 1}^k \sum_{\mathbf{h} \in \mathcal{C}_i}\left(\mathbf{h} - \boldsymbol{\mu}_i\right)\left(\mathbf{h} - \boldsymbol{\mu}_i\right)^\intercal$\;
	$\mathbf{V}\leftarrow \KwEig \left(\mathbf{S}_{w}\right)$\tcc*[r]{$\mathbf{V}$ contains the eigenvectors sorted in ascending order w.r.t. their eigenvalues}
	Using Equation (\ref{eqn:finetune}) to optimize the representation\;
}
\Return{$\mathcal{C}$}
\end{algorithm}


%% file: sec_experiments.tex
\section{Experimental Evaluation}

\subsection{Datasets}

\begin{figure}[htbp]
	\centerline{\includegraphics[width=\columnwidth]{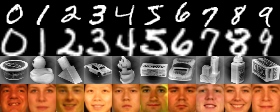}}
	\caption{Some image samples, from top row to bottom row, come from MNIST, USPS, COIL-20 and FRGC datasets, respectively.}
	\label{fig_samples}
\end{figure}

To evaluate the performance and generality of DEKM, we conduct experiments on benchmark datasets and compare with state-of-the-art methods. In order to show that DEKM works well on various datasets, we choose four image datasets that cover domains such as handwritten digits, objects,
and human faces and three text datasets. Table \ref{tab1_ds} provides a brief description for each dataset. And some examples in the image datasets are shown in Figure~\ref{fig_samples}.

MNIST~\cite{lecun1998gradient} is a dataset that consists of 70,000 hand-written grayscale digit images. Each image is of size 28$\times$28 pixels. USPS is a dataset of handwritten grayscale digit images from the USPS postal service, containing 9,298 images of size 16$\times$16 pixels. COIL-20~\cite{nene1996columbia} is a dataset containing 1,440 color images of 20 objects (72 images per object). The objects have a wide variety of complex geometric and reflectance characteristics. The size of each image is resized to 28$\times$28 pixels. FRGC is a human face dataset. Following~\cite{yang2016joint}, we randomly select 20 subjects from the original dataset and collect their 2,462 face images. Similarly, we crop the face regions and resize them into 28$\times$28 pixels. For all the image datasets, each image is normalized by scaling between 0 and 1. REUTERS-10K~\cite{DEC_xie2016unsupervised} contains a random subset of 10,000 samples from REUTERS dataset which has about 810,000 English news stories. REUTERS-10K contains four categories: corporate/industrial, government/social, markets, and economics. The 20 Newsgroups dataset (20NEWS)~\cite{lang1995newsweeder} contains 18,846 documents labeled into 20 different classes, each corresponding to a different topic. Reuters Corpus Volume I (RCV1)~\cite{lewis2004rcv1} contains 804,414 manually categorized newswire stories. Following~\cite{fard2020deep}, we sample from the full RCV1 collection a random subset of 10,000 documents from the largest four categories, denoted by RCV1-10K. For the three text datasets, we represent each document as a tf-idf feature vector on the 2,000 most frequently occurring word stems. And each sample $\mathbf{x}_i$ is normalized so that $\frac{1}{d}\lVert\mathbf{x}_i\rVert_2^2$ is approximately 1, where $d$ is the dimension of the input space.

\begin{table}[htbp]
	\caption{Description of datasets}
	\begin{center}
		\begin{tabular}{|c|c|c|c|}
			\hline
			Dataset                        & Sample \# & Cluster \# & Dimensions \\ \hline
			MNIST                          & 70,000      & 10      & 28$\times$28$\times$1    \\ \hline
			USPS                             & 9,298        & 10      & 16$\times$16$\times$1    \\ \hline
			COIL-20                       & 1,440         & 20      & 28$\times$28$\times$1    \\ \hline
			FRGC                             & 2,462        & 20      & 32$\times$32$\times$3    \\ \hline
			REUTERS-10K              &10,000       &4         &2,000 \\ \hline    
			20NEWS                       &18,846       &20         &2,000 \\ \hline 
			RCV1-10K                      &10,000       &4         &2,000 \\ \hline 
		\end{tabular}
		\label{tab1_ds}
	\end{center}
\end{table}

\subsection{Baseline Methods}
We compare the proposed DEKM with the following methods:
\begin{itemize}
	\item K-means: We perform K-means on the original data.
	\item PCA+K-means: We perform K-means in the space spanned by the first $p$ principal components of data using Principal Component Analysis (PCA), where $p$ is selected to keep  90\% of data variance. 
	\item AE+K-means: We perform K-means in the embedding space of our pretrained convolutional/MLP autoencoder.
	\item DEC~\cite{DEC_xie2016unsupervised}: It uses an MLP autoencoder to find the embedding space and then performs clustering in the embedding space by minimizing the Kullback-Leibler (KL) divergence between the cluster distribution and a target Student’s t-distribution. 
	\item DCEC~\cite{DCEC_guo2017deep}: It replaces the MLP autoencoder in DEC with a convolutional autoencoder.
	\item IDEC~\cite{IDEC_guo2017improved}: It replaces the MLP autoencoder in DEC with a under-complete autoencoder that preserves the local structure of data.
	\item DCN~\cite{DCN_yang2017towards}: It combines the objective function of the MLP autoencoder with that of K-means and alternately optimizes them.
	\item DKM~\cite{fard2020deep}: It considers the K-means objective as a limit of a differentiable function and adopts stochastic gradient descent to jointly optimize representation learning and clustering.
\end{itemize} 

\subsection{Experimental Settings}

Since convolutional neural network (CNN) is good at capturing the semantic visual features of input images, we exploit convolutional autoencoder to find the embedding space for the image datasets. Specifically, we use three convolutional layers followed by a dense layer (embedding layer) in the encoder-to-decoder pathway. The channel numbers of the three convolutional layers are 32, 64, and 128 respectively. The kernel sizes are set to 5$\times$5, 5$\times$5, and 3$\times$3 respectively. The stride of all the convolutional layers is set to two. The number of neurons in the embedding layer is set to the number of clusters of datasets. The decoder is a mirror of the encoder and the output of each layer of the decoder is appropriately zero-padded to match the input size of the corresponding encoder layer. All the intermediate layers of the convolutional autoencoder are activated by ReLU \cite{krizhevsky2012imagenet}. For the text datasets, we use a fully connected multilayer perceptron (MLP) for the backbone of autoencoder. Following the settings in DEC~\cite{DEC_xie2016unsupervised}, the encoder has dimensions of $d$-500-500-2000-10, where $d$ is the dimension of the input data. The decoder is a mirror of the encoder. All the intermediate layers are activated by ReLU. The weights of all the layers are initialized by Xavier approach \cite{glorot2010understanding}. The Adam \cite{kingma2014adam} optimizer is adopted with the initial learning rate  $l=0.001$, ${\beta_1} = 0.9$, ${\beta_2} = 0.999$. 
We stop the clustering process when there are less than 0.1\% of samples that change their clustering assignments between two consecutive iterations. Our code is available at: \textcolor{blue}{\url{https://github.com/spdj2271/DEKM}}. All the experiments are conducted on a machine with one Intel(R) Xeon(R) CPU (2.00GHz) and one Nvidia Tesla P100 GPU.


\subsection{Evaluation Metric}

To evaluate the clustering methods, we adopt two standard evaluation metrics: normalized mutual information (NMI)~\cite{vinh2010information} and unsupervised clustering accuracy (ACC) \cite{xu2003document}. Both the NMI and ACC values are in the range $\left[ 0, 1\right] $. The higher the values, the better the clustering results. NMI is an information-theoretic measure, which calculates the normalized measure of similarity between the ground-truth labels and the obtained cluster assignments. NMI is defined as follows:

\begin{equation}
	\mbox{NMI} = \frac{2\times I(\mathcal{G};\mathcal{C})}{H(\mathcal{G})+H(\mathcal{C})}
\end{equation}
where $\mathcal{G}$ is the ground-truth, $\mathcal{C}$ is the cluster assignments, $I$ denotes mutual information, and $H$ denotes entropy.

ACC measures the proportion of samples whose cluster assignments can be correctly mapped to the ground-truth labels. ACC is defined as follows:
\begin{equation}
	\mbox{ACC} = \max_m \frac{\sum_{i = 1}^n \mathds{1}\left\{ \mathbf{g}_i = m\left( \mathbf{c}_i \right) \right\} }{n}
\end{equation}
where $\mathbf{g}_i$ is the ground-truth label of the $i$-th data point, $\mathbf{c}_i$ is the cluster assignment of the $i$-th data point, $m$ ranges over all possible one-to-one mappings between ground-truth labels and cluster assignments. The mapping is based on the Hungarian algorithm \cite{kuhn2005hungarian}. 

\begin{table*}[!htb]
	\caption{Clustering results of different algorithms in terms of the unsupervised clustering accuracy (ACC\%) and normalized mutual information (NMI\%). The results marked by $^*$ come from the original papers. $-$ denotes that the result is not available. DEKM\_F means that DEKM uses the full-batch updating strategy.}
	\begin{center}
		\resizebox{\textwidth}{!}{
		\begin{tabular}{|l|l|l|l|l|l|l|l|l|l|l|l|l|l|l|}
			\hline
			\multirow{2}{*}{Method} & \multicolumn{2}{c|}{MNIST}      & \multicolumn{2}{c|}{USPS}       & \multicolumn{2}{c|}{COIL-20}    & \multicolumn{2}{c|}{FRGC}    & \multicolumn{2}{c|}{REUTERS-10K} & \multicolumn{2}{c|}{20NEWS} & \multicolumn{2}{c|}{RCV1-10K}   \\ \cline{2-15} 
			& ACC                       &NMI                 &ACC               &NMI                  &ACC             &NMI            &ACC             &NMI            & ACC          & NMI        & ACC          & NMI        & ACC          & NMI      \\ \hline
			K-means                                                       &53.40                      &50.00             &66.81             &62.62               &25.28          &67.17          &12.63          &18.27          &54.05         &41.91       &21.59         &19.70      &51.96          &31.35\\ \hline
			PCA+K-means                                           &53.13                      &49.91              &54.20             &60.96              &26.53          &65.34          &12.55          &15.19          &53.97         &41.35      &21.75         &20.46      &51.91          &31.39\\ \hline
			AE+K-means                               &85.47                     &80.53             &74.84             &74.16              &68.82           &79.56          &33.31          &42.81          &70.52          &49.02      &40.24        &37.60      &64.11         &41.12\\ \hline
			DEC    &84.30$^*$           &83.72$^*$     &73.68$^*$     &75.29$^*$     &65.42$^*$   &80.49$^*$ &37.10$^*$  &44.60$^*$  &72.17$^*$    &54.87          &35.65            &43.79          &66.81           &45.24        \\ \hline
			DCEC            &87.16                     &85.92              &78.35              &81.34              &71.88          &\textbf{82.24}            &33.40        &41.50    &72.17    &54.87          &35.65            &43.79          &66.81           &45.24        \\ \hline
			IDEC        &88.06$^*$           &86.72$^*$     &76.05$^*$     &78.46$^*$      &$-$             &$-$                &$-$              &$-$           &75.64$^*$         &49.81$^*$      &40.50        &38.20       &59.50      &34.70   \\ \hline
			DCN         &84.00                   &80.00               &67.00             &67.00               &\textbf{74.00} &81.00       &$-$              &$-$           &$-$            &$-$           &44.00$^*$        &\textbf{48.00}$^*$      &56.70       &31.60       \\ \hline
			DKM                         &84.00$^*$            &79.60$^*$      &75.70$^*$    &77.60$^*$     &$-$             & $-$             &$-$              &$-$             &$-$              &$-$          &\textbf{51.20}$^*$         &46.70$^*$       &58.30$^*$       &33.10$^*$     \\ \hline
			DEKM                                                             &\textbf{95.75}       & \textbf{91.06} &\textbf{79.75}   &\textbf{82.23}   &69.03   &80.06     &\textbf{38.59}    &\textbf{50.78}     &76.28&59.06&41.08&40.27&\textbf{67.15}&\textbf{46.18}     \\ \hline
			DEKM\_F                                                   &94.23                       &90.54                &78.87          &80.51          &72.62          &81.97          &37.29          &49.32    &\textbf{76.42}&\textbf{59.99}&41.20&43.76&62.12&35.98      \\ \hline
		\end{tabular}}
		\label{tab_all}
	\end{center}
\end{table*}

\subsection{Clustering Results}

Due to randomizations of the weights of autoencoder and the centroids of K-means, we run DEKM on each dataset for three times and report the average results. Table~\ref{tab_all} reports the clustering performances of different methods on the benchmark datasets in terms of normalized mutual information (NMI) and unsupervised clustering accuracy (ACC). For the comparison methods, we present the reported results from their original papers if they are available (marked by $^*$ in Table~\ref{tab_all}). For unreported results on specific datasets, we run the released code mentioned in the original papers. When the released code is not available or running it is not practical, we put dash marks ($-$) in Table~\ref{tab_all}. 

We can see from Table~\ref{tab_all} that DEKM outperforms the comparison methods on most of the datasets. DEKM achieves competitive results to DCEC on COIL-20 in terms of NMI. DEKM outperforms all the comparison methods with a large margin on MNIST, 20NEWS and RCV1-10K. DEKM\_F uses the full-batch updating strategy to optimize the representation. Compared with DEKM, DEKM\_F performs slightly worse on four datasets MNIST, USPS, FRGC, and RCV1-10K. As can be seen, all the deep clustering methods perform much better than the traditional shallow clustering methods (i.e., K-means and K-means+PCA). This suggests that the embedding space generated by autoencoder is more advantageous for clustering. The performance gap between DEKM and AE+K-means is large, which means our representation optimization strategy is promising. Both DEKM and DCEC use convolutional autoencoders to find the embedding space for the image datasets. The performance gap between DEKM and DCEC reflects the effect of different representation optimization strategy. The representation optimization strategy of DEKM is superior to that of DCEC. Note that DCEC replaces the MLP autoencoder in DEC with a convolutional autoencoder. For the text datasets, DCEC using the MLP autoencoder is equivalent to DEC. Compared with DCEC on the text datasets, we can see that DEKM also performs better. Thus, the representation optimization strategy of DEKM works in different scenarios, making DEKM a universal clustering framework.

\subsection{Representation Optimization Strategy }

\begin{figure}[!htb]
	\centering
	\includegraphics[width=0.45\textwidth]{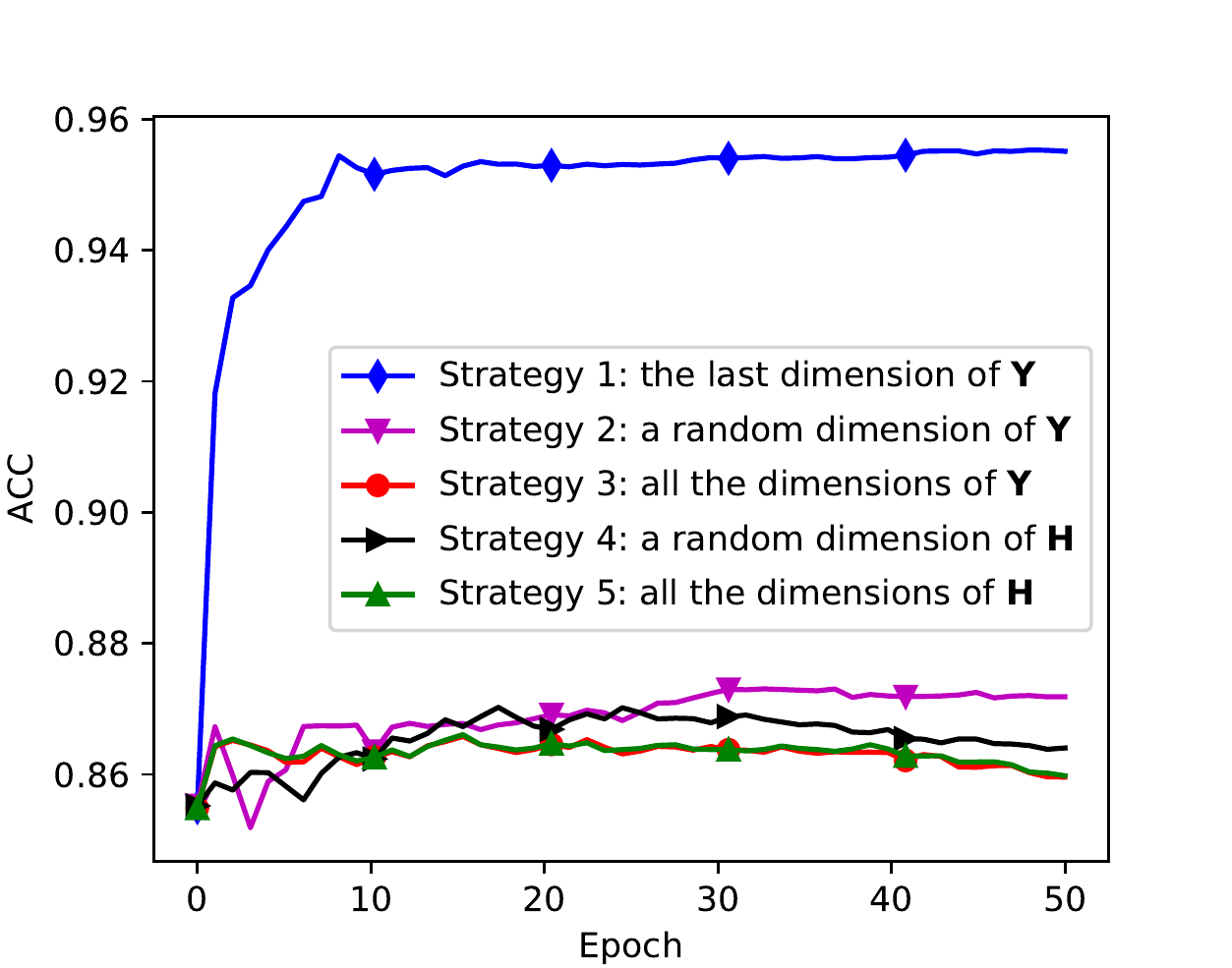}
	\caption{The clustering performances of the different representation optimization strategies on MNIST dataset.}
	\label{fig_ACC}
\end{figure}

We examine the effects of several representation optimization strategies on the MNIST dataset. Specifically, we compare the following strategies: 1. reducing the entropy of the last dimension of $\mathbf{Y}$, 2. reducing the entropy of a random dimension of $\mathbf{Y}$, and 3. reducing the entropy of all the dimensions of $\mathbf{Y}$. We also compare with another two strategies: 4. reducing the entropy of a random dimension of $\mathbf{H}$, and 5. reducing the entropy of all the dimensions of $\mathbf{H}$. Note that all these strategies use the mini-batch updating strategy. Figure~\ref{fig_ACC} shows the comparison results. We can see that the first strategy of entropy reduction in the last dimension of $\mathbf{Y}$ is the best. It outperforms the other four strategies with a large margin. Strategy 2 performs better than strategy 4. Strategy 3 performs similarly to strategy 5.

\subsection{Embedding Space Comparison}
\begin{figure}[!htb]
	\hspace*{\fill}
	\centering	
	\subfigure[Raw data + PCA]{\includegraphics[width=0.2\textwidth]{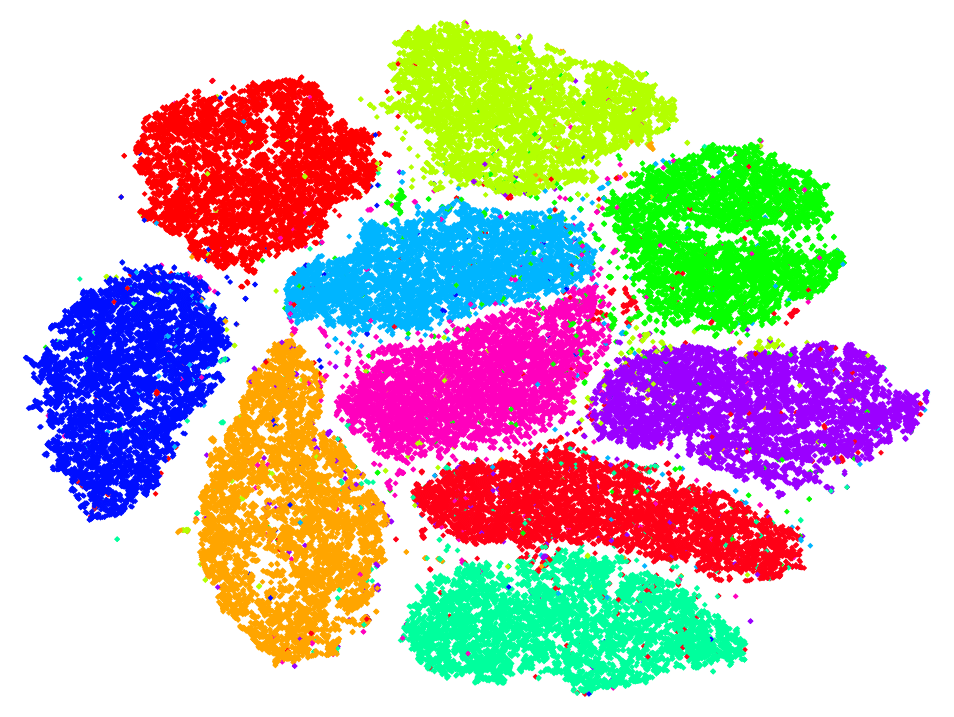}}
	\hfill
	\subfigure[AE]{\includegraphics[width=0.2\textwidth]{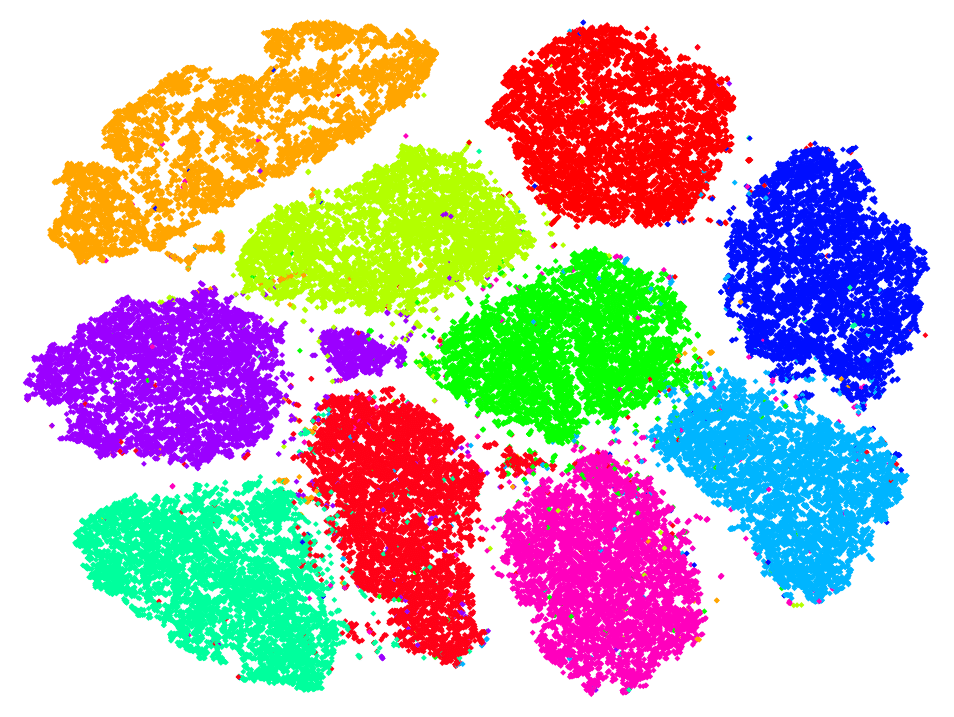}}
	\hspace*{\fill}
	
	\hspace*{\fill}
	\centering	
	\subfigure[DEC]{\includegraphics[width=0.2\textwidth]{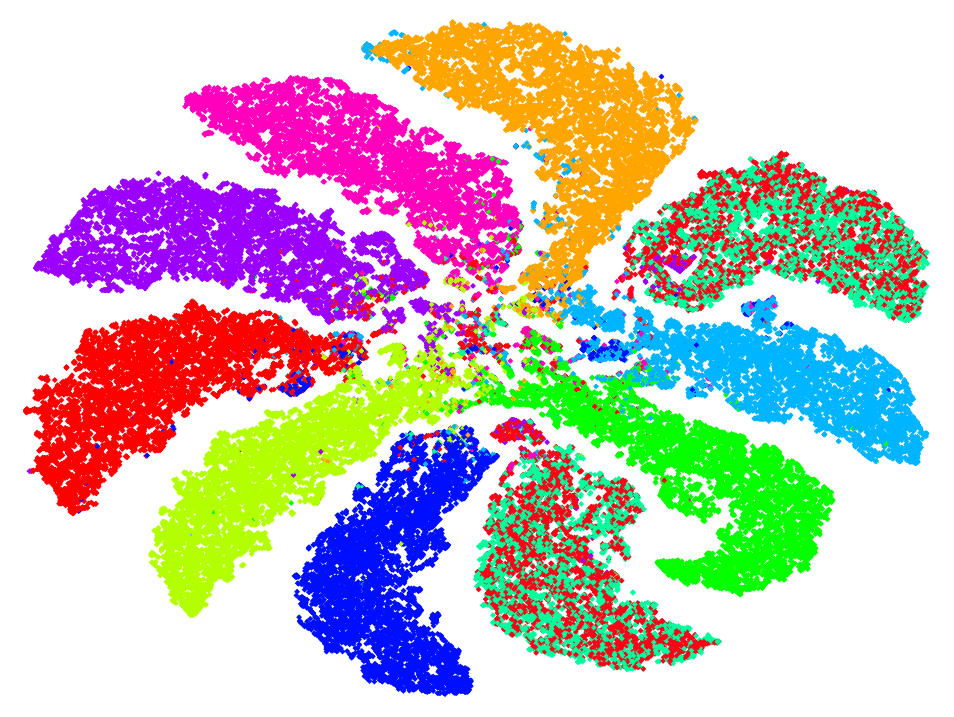}}
	\hfill
	\subfigure[DEKM]{\includegraphics[width=0.2\textwidth]{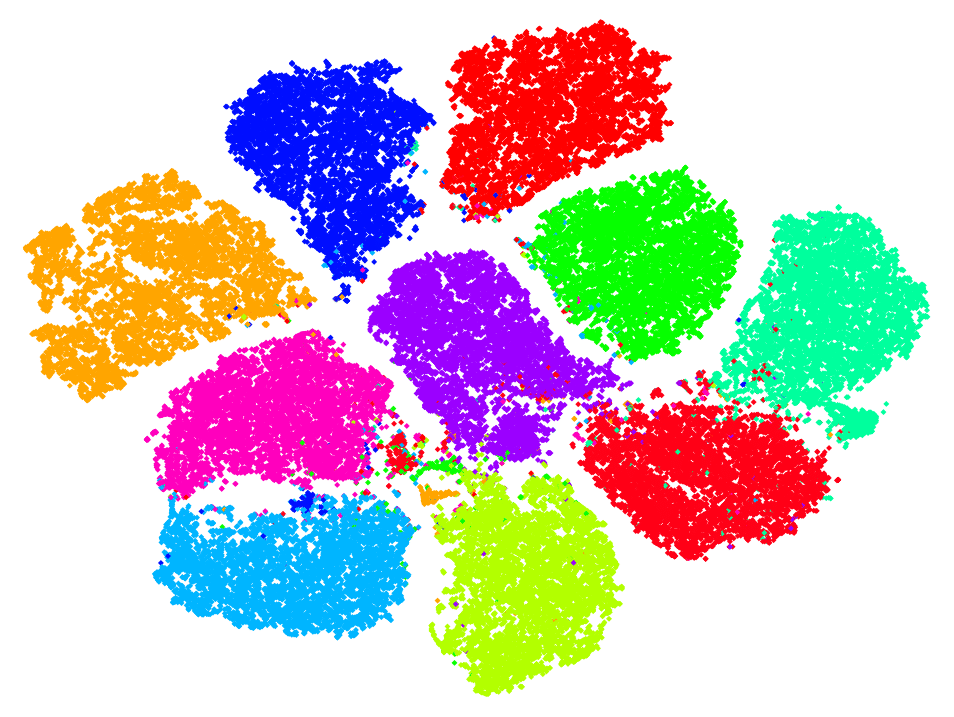}}
	\hspace*{\fill}
	\caption{The t-SNE visualization of the embedding spaces of different algorithms on MNIST dataset.}
	\label{fig_em}
\end{figure}

Figure~\ref{fig_em} illustrates the t-SNE~\cite{van2008visualizing} visualization of the embedding spaces of different algorithms on MNIST dataset. Figure~\ref{fig_em}(a) demonstrates the embedding space of PCA. Figure~\ref{fig_em}(b) illustrates the embedding space of a convolutional autoencoder, which is the initial embedding space for DEKM. Figure~\ref{fig_em}(c) shows the embedding space of DEC. And Figure~\ref{fig_em}(d) shows the embedding space of DEKM. Note that all these embedding spaces are used to get the clustering results in Table~\ref{tab_all}. Compared with the clusters in the initial embedding space (as shown in Figure~\ref{fig_em}(b)) of the convolutional autoencoder, the clusters in the embedding space (as shown in Figure~\ref{fig_em}(d)) of DEKM are more focused and isotropic, which are good for K-means. The two clusters in the embedding space of DEC are mixed, which leads to a lower performance compared with DEKM. The clusters in the embedding space of PCA are not isotropic Gaussian clusters, which is the reason why K-means does not perform well.

%
%

%% file: sec_conclusion.tex
\section{Conclusion}

In this paper, we have proposed a new deep clustering method called DEKM that alternately learns the deep embedding space and finds clusters inside. First, we train an autoencoder to generate an embedding space. Then in this embedding space we find clusters by K-means. We eigendecompose the within-class scatter matrix of K-means to get an orthonormal transformation matrix, which is further used to transform the embedding space to a new space that reveals the cluster-structure information. Each row vector of the orthonormal transformation matrix is the eigenvector of the within-class scatter matrix. The eigenvalue indicates the importance of the corresponding eigenvector's contribution to the cluster-structure information in the new space. Finally, we propose a greedy method to optimize the representation to increase the cluster-structure information in the new space. We optimize DEKM in an alternating way. Experimental results show that DEKM achieves superior performances compared to the baselines. In the future, we would like to extend DEKM to find non-redundant clusters in multiple independent subspaces.